\documentclass[10pt, a4paper]{article}
\usepackage{lrec2022} %
\usepackage{multibib}
\newcites{languageresource}{Language Resources}
\usepackage{graphicx}
\usepackage{tabularx}
\usepackage{soul}
\usepackage{titlesec}
\titleformat{\section}{\normalfont\large\bfseries\center}{\thesection.}{1em}{}
\titleformat{\subsection}{\normalfont\SmallTitleFont\bfseries\raggedright}{\thesubsection.}{1em}{}
\titleformat{\subsubsection}{\normalfont\normalsize\bfseries\raggedright}{\thesubsubsection.}{1em}{}
\renewcommand\thesection{\arabic{section}}
\renewcommand\thesubsection{\thesection.\arabic{subsection}}
\renewcommand\thesubsubsection{\thesubsection.\arabic{subsubsection}}

\usepackage{epstopdf}
\usepackage[utf8]{inputenc}

\usepackage{hyperref}
\usepackage{xstring}

\usepackage{color}

\title{Collecting Visually-Grounded Dialogue with A Game Of Sorts}

\name{Bram Willemsen, Dmytro Kalpakchi, Gabriel Skantze} 

\address{Division of Speech, Music and Hearing, KTH Royal Institute of Technology \\
         Stockholm, Sweden \\
         \{bramw, dmytroka, skantze\}@kth.se\\}

\abstract{
An idealized, though simplistic, view of the referring expression production and grounding process in (situated) dialogue assumes that a speaker must merely appropriately specify their expression so that the target referent may be successfully identified by the addressee. However, referring in conversation is a collaborative process that cannot be aptly characterized as an exchange of minimally-specified referring expressions. Concerns have been raised regarding assumptions made by prior work on visually-grounded dialogue that reveal an oversimplified view of conversation and the referential process. We address these concerns by introducing a collaborative image ranking task, a grounded agreement game we call “A Game Of Sorts”. In our game, players are tasked with reaching agreement on how to rank a set of images given some sorting criterion through a largely unrestricted, role-symmetric dialogue. By putting emphasis on the argumentation in this mixed-initiative interaction, we collect discussions that involve the collaborative referential process. We describe results of a small-scale data collection experiment with the proposed task. All discussed materials, which includes the collected data, the codebase, and a containerized version of the application, are publicly available.
 \\ \newline  }

\begin{document}

\maketitleabstract
\setcounter{figure}{1}

\section{Introduction}
\label{section-intro}
Visually-grounded dialogues are conversations in which entities in a co-observed visual context are referenced. 
In order for an addressee to successfully ground a referring expression, the description of the referent should be appropriately specified. 
If a speaker were to abide by the Gricean Maxim of Quantity \cite{Grice1975LogicConversation}, we would expect them to produce a referring expression containing precisely the content necessary for the addressee to identify the target referent.
Instead, in addition to such minimally-specified referring expressions, we commonly observe those that are over- and underspecified, meaning they contain more or less information than is strictly necessary to correctly ground the description \cite{Arts2011OverspecificationIdentification,Koolen2011FactorsDescriptions,Gatt2013ProductionDiscrimination,Rubio-Fernandez2019OverinformativeQuantity}. 
Moreover, as dialogue is inherently an interactive process, producing a description for a referent is often a collaborative effort, rather than a unilateral transfer of well-formed, unambiguous referring expressions \cite{Clark1986ReferringProcess}. 
Notably, participants in a conversation will iteratively refine and simplify their descriptions when repeatedly addressing the same referent. 
Over time, the conversational partners may form so-called conceptual pacts, as they come to an (implicit) agreement on a shared conceptualization of a referent \cite{Clark1986ReferringProcess}. 
Established pacts are said to be part of the common ground \cite{Stalnaker1978Assertion,Clark1986ReferringProcess,Brennan1996ConceptualConversation,Clark1996UsingLanguage}, i.e., that which is believed by the conversational partners to be mutual knowledge. 
However, even seemingly stable pacts are not immutable and may eventually be refashioned or abandoned altogether \cite{Bangerter2020LexicalTask}.

The production and resolution of grounded references in conversation is complex and dynamic. If we want to model this process, we require data that is representative and, thus, reflects that complexity to a satisfactory degree. 
However, concerns have been raised regarding the oversimplification of visually-grounded dialogue tasks and resulting data \cite{Haber2020TheDialogue,Ilinykh2019MeetUpEnvironment,Schlangen2019GroundedSettings}.
Prior work in this area often restricts the dynamics of conversation and prescribes constraints in the interest of experimental control. 
While limiting the scope of the problem makes modeling and evaluating the task more manageable, it often comes at the cost of the task and, with it, the collected data representing visually-grounded dialogue in a strikingly limited sense; the issue being that the imposed constraints are reflected in the data and the principles induced from the data may not generalize beyond the task from which they were derived.
For instance, claims made based on observations from fixed-initiative interactions, such as those from role-asymmetric multi-turn visual question-answering tasks \cite{Das2017,DeVries2017GuessWhatDialogue}, are not necessarily extendable to mixed-initiative dialogues.
Other principal considerations include the permitted language use, as tasks that restrict the content of the messages are bound to artificially reduce the lexical diversity of the collected data. Similarly, restrictions such as a time, turn, word, or character limit, when particularly constraining, will inevitably influence the way in which participants communicate with each other.
By purposefully avoiding the role asymmetry common to prior work, providing an objective that incorporates realistic stimuli and for which the participants are jointly and equally responsible, \newcite{Ilinykh2019MeetUpEnvironment} and \newcite{Haber2020TheDialogue} manage to more expressly capture dialogue phenomena that are expected in conversations in which participants reference a visual modality. 
Even so, the level of reasoning over the visual information required for task success does not go beyond what is necessary for the production of appropriately-specified descriptions.
This is the result of these tasks \cite{Haber2020TheDialogue,Ilinykh2019MeetUpEnvironment} effectively being cooperative games with imperfect information, where the main objective is for each player to share the information available to them but that may be hidden from the other player, so as to create a game with perfect information through conversation.

With the intention of providing a less restrictive resource for modeling and evaluating visually-grounded dialogue models, we introduce a collaborative image ranking task we call \textbf{A Game Of Sorts}.
The task is presented as a game in which players are challenged to come to an agreement through largely unrestricted, role-symmetric dialogue on how to rank a set of images given some sorting criterion. 
We adopt the notion of information asymmetry leveraged by prior work \cite{Haber2020TheDialogue,Ilinykh2019MeetUpEnvironment,Udagawa2019AContext} to force descriptions about the content of the stimuli, making the task a game with imperfect information.
However, unlike prior work, we make resolving the information asymmetry a secondary objective; the primary objective for the players is to arrive at a ranking that is satisfactory for all parties involved in the game. 
Agents that pay heed to the primary objective do not only need to generate and ground referring expressions, but will also have to reason about how the sorting criteria relate to the visual stimuli. They are thereby encouraged to argue their point of view, providing explicit motivations, whilst also having to understand the motivations of others. 
By emphasizing the importance of argumentation, we aim to 
collect data with a rich mixture of dialogue acts, that is lexically diverse, and that to a greater degree captures dialogue phenomena underrepresented or absent in prior work.
Note that, although the ranking of images remains the overarching undertaking throughout the game, the sorting criterion changes from round to round, meaning that the players will have to adapt their strategies accordingly.

We expect that the problem we propose here is more challenging to solve end-to-end than those posed by more restrictive tasks, such as the aforementioned task of multi-turn visual question-answering \cite{Das2017,DeVries2017GuessWhatDialogue} even when it emphasizes the need for discourse memory \cite{Agarwal2020HistoryIt}. This is due to the mixed-initiative and generally more unrestricted nature of our game, the involvement of argumentation, and the reliance on commonsense reasoning. This makes data collected with our game a potentially challenging test set for downstream tasks such as coreference resolution, referring expression generation and comprehension, and dialogue act classification. %
Additionally, our game facilitates the study of the effect of distractors on content selection and lexical choice in the collaborative referring expression production process \cite{Zarrie2018DecodingGeneration}. 

Our main contributions are as follows:
\begin{itemize}
    \item We describe a new grounded agreement game \cite{Schlangen2019GroundedSettings} we call \textbf{A Game Of Sorts},
    and argue for its use in modeling and evaluating multimodal dialogue models in terms of their referring expression generation and grounding capabilities;
    \item We report on a small-scale data collection experiment using the task with its proposed setup and provide an analysis of the collected data, contrasting our data with that of tasks from closely-related prior work \cite{Haber2020TheDialogue,Ilinykh2019MeetUpEnvironment};
    \item We make all materials, i.e., the collected data, the (documented) codebase, and a containerized version of the application, publicly available to facilitate the extension and reproduction of our work\footnote{\url{https://github.com/willemsenbram/a-game-of-sorts}, \href{https://doi.org/10.5281/zenodo.6489019}{doi:10.5281/zenodo.6489019}}.
\end{itemize}

\section{Related Work}
There exists a large body of work on the collection of referring expressions in visually-grounded dialogue \cite{Tokunaga2012TheDialogues,Zarrie2016PentoRef:Dialogues,Shore2018,Haber2020TheDialogue,Ilinykh2019MeetUpEnvironment,Udagawa2019AContext,Kottur2021SIMMCConversations}. We focus on two relatively recent works in particular, these being the \textit{MeetUp!} corpus \cite{Ilinykh2019MeetUpEnvironment} and the \textit{PhotoBook} dataset \cite{Haber2020TheDialogue}, as we believe them to be most similar to the work presented in this paper.

MeetUp! and PhotoBook can be considered \textit{grounded agreement games} \cite{Schlangen2019GroundedSettings}, as both tasks are focused on a (mostly) unrestricted, role-symmetric dialogue through which players have to come to an agreement on an answer to a given question using the (visual) information available to each participant.

The MeetUp! task is presented as a cooperative game in which two players navigate a virtual environment, represented by static images of real-world scenes, with the goal of meeting up in the same location. 
Navigation happens under partial knowledge, as the two players cannot see each other's perspective. This forces them to describe their surroundings, i.e., the content of the image, in order to understand whether they have successfully managed to navigate to the same location.

The PhotoBook task was similarly introduced as a two-player cooperative game. Each player is shown a number of pictures: some of these are shown to both players, while others are shown to either player. The goal of the game is for both players to find out which images they do and which images they do not have in common. Ergo, PhotoBook, similar to MeetUp!, is a game with imperfect information. The fact that the game is played over several rounds and a number of images recurs over the course of the interaction makes coreferences as well as the forming of conceptual pacts more likely. %

Although both tasks succeed in capturing various dialogue phenomena previously underrepresented or entirely absent in prior work, the primary objective for each boils down to image matching: the task is centered around reaching a game with perfect information through conversation. Virtually no additional reasoning is required, making the most efficient form of play one that involves little to no dialogue but instead devolves into an exchange of overspecified referring expressions.
We propose to further increase the likelihood of productive conversations between players that abide by the cooperative principle \cite{Grice1975LogicConversation} by introducing argumentation, making resolving partial knowledge a secondary objective.
We will contrast data collected using our proposed task with that of MeetUp! and PhotoBook.

\section{A Game Of Sorts}
A Game Of Sorts is an image ranking task framed as a two-player, cooperative game. Participants in this game are presented with a set of images and a criterion by which to sort them. 

\subsection{Gameboard}
The images on the gameboard are displayed in a grid, such as shown in Figure \ref{figure-interface} (see Appendix). Both participants see the same images, though
their position on the grid is randomized separately for each player.  
This forces a degree of \textit{imperfect information} as players will not be able to refer to images using spatial relations but must instead describe them by their content.
The image sets are constructed so that each image has a %
number of semantically-similar %
counterparts %
in order to increase the likelihood of non-trivial referring expressions. 

\subsection{Sorting Criteria}
The game is played over multiple rounds with a recurring set of images, forcing repeated references.
However, each round has a different sorting criterion by which the players will have to rank the images. 
The sorting criterion does not necessarily 
need to hint at an objective resolution. In fact, in order to spark a discussion that could increase the length of the conversation it may be beneficial
if the criterion steers towards a somewhat contentious topic of conversation or is otherwise open to interpretation. 

\subsection{Communication between Players}
Players communicate with each other by exchanging text messages.
The interaction is role-symmetric, as the participants are not restricted by
predefined roles. Messages are similarly unrestricted, as we do not impose a character limit nor prescribe the content of an utterance, allowing references to one or multiple images, or the absence of referring expressions altogether. 
Players are encouraged to explicitly motivate their propositions and discuss their thoughts at length whenever appropriate, which should increase the likelihood of a wide range of dialogue acts manifesting over the course of the interaction.

\subsection{Self-Annotation}
In order to aid (manual) annotation efforts, players are required to explicitly indicate whether or not their message contains a referring expression.
In the event that their message contains a reference to one ore more images, the participant selects all intended referents by clicking the corresponding images on the grid, prior to sending the message.
In case the message contains no reference to an image, the participant is asked to click a designated button to indicate as much instead.
By enforcing this means of \textit{self-annotation} we ensure underspecified referring expressions can be resolved and mapped to their respective target referents, post hoc. Note that the receiving player does not see which images (if any at all) were selected by the player sending the message.

\subsection{Locking Images}
When players have come to an agreement on how to rank one or more of the images, they will have to indicate their choice by \textit{locking in} the image or images, one at a time. An image is locked when a player selects an image and then clicks the \textit{lock} button. Each player does this individually, without being able to see which image was locked by the other player. Only when both players have locked in an image will they receive feedback on their action. In the event that both players locked in the same image it will be successfully ranked, which is then visually indicated. However, if each player locked in a different image the locked image will be unlocked and deselected and both players informed that they are not aligned on the same image.
Once an image has been successfully ranked, the choice is final and players cannot undo or otherwise change this action. The round ends when all images on the grid have been ranked successfully. 

\subsection{Grounded Agreement Game}
Formally, A Game Of Sorts fits the definition of a grounded agreement game \cite{Schlangen2019GroundedSettings}: 
two participants $P = \{P_{1}, P_{2}\}$ are tasked by a third party, moderator $M$, to sort a set of images $I$ using criterion $C$. 
However, rather than the game ending after a singular agreed upon answer, a round is over when the number of agreed upon answers in the set of all answers $A$ is equal to the number of elements in the set $I$, so that %
$|A| = |I|$.
Moreover, cooperation happens under partial knowledge, as some information regarding $I$ is dispersed (i.e., each participant sees the same images, but their order is randomized and some actions in relation to $I$ taken by one player are not immediately visible to the other), making this a game with imperfect information. Only when both players have locked an image will they receive feedback on their action.

\subsection{Guaranteeing Repeated References}
Note that it is possible to reduce the number of images to be ranked, such that $|A| < |I|$, but still guarantee repeated references. We can calculate the minimum number of rounds needed to guarantee at least one repeated reference as
$$ R = \bigg\lfloor \frac{|I|}{|A|} \bigg\rfloor + 1 $$ 
where $|I|$ is the total number of images on the grid, and $|A|$ is the total number of images to be ranked each round, so as long as $A \neq \emptyset$.

\subsection{Basic Principles}
Although players are always tasked with ranking images according to some sorting criterion $C$, $C$ changes from round to round, requiring the participants to adapt to a dynamic task context. For effective collaboration, we expect each participant to be able to assess the quality of propositions made by another player as well as make reasonable propositions of their own. The level of reasoning involved for a participant to relate $C$ to $I$ goes beyond the generation of unambiguous referring expressions. It requires the participant to understand each element of the compound scene $I$ and how $C$ affects the interpretation of each individual element. Each player performs an implicit ranking for $I$ based on $C$, which allows them to evaluate whether to accept or reject proposals by the other player, as propositions that align with their preliminary ranking can be considered reasonable for acceptance, while those that do not will require further discussion or are rejected instead. When it becomes clear that ranking strategies between $P_{1}$ and $P_{2}$ diverge is when motivated reasoning becomes especially relevant. The challenge then is to understand whether proposals can be considered reasonable given additional explanation, which will likely lead to acceptance, or whether another proposition is more reasonable still, leading to rejection and the need for a motivated counter-proposal.

\begin{table*}[!t]
\footnotesize
\begin{center}
\begin{tabularx}{2\columnwidth}{llll}

      \hline
       & MeetUp!$^a$ & PhotoBook$^b$ & \textbf{A Game Of Sorts} (ours) \\
      \hline
       \textit{\#} Dialogues & 430 & 2,506 & 15 \\

       \textit{\#} Utterances & 5,695  & 164,296 & 1,800  \\
       
       \textit{\#} Sentences & 6,020  & 172,550 & 2,274  \\

       \textit{\#} Tokens & 31,431  & 1,038,353 & 19,811  \\
       
       \textit{\#} Types$^c$ & 1,948 & 10,724 & 1,720  \\
       
       \hline
       
       Average Dialogue Length (Utterances)$^d$ & 13.24 \textit{(6.54)} & 65.67 \textit{(14.90)} & 120.00 \textit{(19.04)} \\
       
       Average Dialogue Length (Sentences)$^d$ & 14.00 \textit{(6.80)} & 68.96 \textit{(16.85)} & 151.60 \textit{(28.46)} \\

       Average Dialogue Length (Tokens)$^d$ & 73.10 \textit{(41.80)} & 415.01 \textit{(157.63)} & 1,320.73 \textit{(436.64)} \\
       
       Average Utterance Length (Tokens)$^d$ & 5.52 \textit{(4.53)} & 6.32 \textit{(5.12)} & 11.01 \textit{(9.53)} \\
       
       Average Sentence Length (Tokens)$^d$ & 5.22 \textit{(3.86)} & 6.02 \textit{(4.79)} & 8.71 \textit{(6.81)} \\
       
       \hline

\end{tabularx}
\caption{Descriptive statistics for \textbf{A Game Of Sorts} and related work. $^a$\protect\newcite{Ilinykh2019MeetUpEnvironment}. $^b$\protect\newcite{Haber2020TheDialogue}. $^c$Number of unique tokens (vocabulary). $^d$Standard deviation in brackets.} %
\label{table-descriptives}
 \end{center}
\end{table*}

\section{Method}
\label{sectionmethod}
To characterize the data collected with the proposed task, we conducted a small-scale data collection experiment, the setup of which is described %
in this section.

\subsection{Participants}
For the dataset reported in this paper we collected contributions using a convenience sample of 14 participants (7 female, 6 male, 1 non-binary; $M_{age} = 28.00$ years, $SD_{age} = 5.54$ years, $min_{age} = 22$ years, $max_{age} = 42$ years). Participants reported a wide range of first languages, including Arabic, Chinese, Dutch, and Telugu. Although our sample includes just one native English speaker, the average self-reported English language proficiency, measured on a 5-point Likert scale, was high at $4.43$ ($SD = 0.73$). 
Most participants (8) played more than one game ($M = 2.14$, $SD = 1.36$).  %
Participants were financially compensated for their contributions.

\subsection{Materials}
All materials described are available at \url{https://github.com/willemsenbram/a-game-of-sorts}, \href{https://doi.org/10.5281/zenodo.6489019}{doi:10.5281/zenodo.6489019}.

\subsubsection{Images}
The visual stimulus for each game was a methodically-selected set of nine images. The main %
subject of each image was an entity from a shared %
category. 
Each image was chosen so that there were exactly two other images 
with which they had one or more (visual) attributes %
in common that were not shared with 
the other six images in the set.
This was to discourage the use of trivial referring expressions and to allow for the study of referring expressions under the presence of various combinations of distractors with differing degrees of similarity to the referent.
Note that certain images were not considered for selection, for example because they were clearly edited, grayscale, or had watermarks present.

In order to be able to study the effect of the image category on the referring expression production process and the dialogue in general, we constructed image sets for five different image categories. The chosen image categories were dogs (animal), mobile phones (electronic device), cars (vehicle), pastries (food), and paintings (art). Images of dogs were taken from the Stanford Dogs dataset \cite{Khosla2011NovelDogs}, which itself is a subset of the ImageNet database \cite{JiaDeng2009ImageNet:Database}. For mobile phones, cars, and pastries, we selected images from Open Images V6 \cite{Kuznetsova2020TheScale}. For images of paintings we used the WikiArt dataset as introduced in \newcite{Saleh2015Large-scaleFeature}.
We collected data for five games, with each game focused on a single category represented by a set of nine images, meaning 45 images in total.

\subsubsection{Sorting Criteria}
Our main concern for this data collection was to generate a discussion between the participants about the visual stimuli that would, aside from the production of referring expressions, naturally lead to a variety of dialogue acts. For this reason, the sorting criteria were created in such a way that devising a ranking strategy demanded a level of reasoning that required some creative thinking from each participant 
as there was no obvious, correct answer (such as a ranking of different mammals in descending order in terms of their average mass), nor was it entirely arbitrary or based of deeply-rooted or innate personal preferences (such as a ranking of individual family members in descending order in terms of the strength of their relationship to the player). The challenge was to, given a set of images, find a balance between scenarios that were thought-provoking, yet possible for players to reach an agreement on after some discussion.

\subsubsection{Questionnaire}
At the end of each game, participants were presented with a self-administered questionnaire. The questions concerned basic demographic information (i.e., age, gender identity, country of origin, native language), English language proficiency, visual acuity, overall experience with the game, and a construct of  %
partner satisfaction adopted from \newcite{Haber2020TheDialogue}. 

\subsection{Procedure}
Prior to the experiment, each participant was sent an e-mail which included some basic information about the game, the compensation they would receive for their contribution, and a unique URL of a personal page through which could schedule their participation. To ensure participants were unaware of the identity of the person they would be paired with, they were instructed not to coordinate participation outside the platform. They were asked to watch a short instructional video as well as read through the written rules prior to the start of their first game. 
Each pair of participants played through four rounds (the order of which was randomized) of a pseudo-randomly assigned game, after which they were presented with the post-game questionnaire. They were prevented from completing any game more than once, meaning each participant was able to play at most five games.

\section{Results}
\label{section-results}
We report the results from the data collection experiment as described in Section \ref{sectionmethod}, providing %
analysis of the dialogues and contrasting the dataset collected for our task with datasets of prior work. A dialogue excerpt can be found in the Appendix.

\subsection{Descriptive Statistics}
In total, we collected 15 interactions in which the assigned game was successfully completed; three interactions for each of the five games. The average time on task was $52$ minutes and $10$ %
seconds ($SD = 11$ minutes, $22$ seconds). Descriptives to characterize the collected data are provided in Table \ref{table-descriptives}. Also shown are the same statistics for MeetUp! \cite{Ilinykh2019MeetUpEnvironment} and PhotoBook \cite{Haber2020TheDialogue}.

Comparing our data to that of MeetUp! and PhotoBook, we found that, on average, dialogues collected with our task were significantly longer in terms of the number of messages exchanged between participants, as well as the number of sentences and tokens. Furthermore, we found that the average length of utterances, calculated as the number of tokens in an utterance averaged over all messages, was similarly longer. We found a similar result even when messages were segmented into sentences, %
although that difference was less pronounced.

\subsection{Lexical Diversity}
As a measure of lexical diversity, we computed the moving-average type-token ratio (MATTR, \newcite{Covington2010CuttingMATTR}). The standard type-token ratio (TTR) for a text is calculated by dividing the number of types (unique tokens) by the total number of tokens, and as such is heavily influenced by the length of the text. 
If we want to compare numbers across corpora, we need to somehow account for differences in size. \newcite{Covington2010CuttingMATTR} proposed MATTR as an alternative to TTR that is unaffected by text length. By calculating the TTR along a sliding window of a fixed size and averaging all obtained ratios we get the MATTR for a given text. 
To further address differences between the corpora and counteract the potential for an order effect, we computed the average MATTR over multiple ($N = 1,000$) randomly-drawn samples. To ensure scores were not affected by different interpretations of size with respect to these datasets, we fixed the sample size along four dimensions, namely the number of dialogues ($N = 10$), utterances ($N = 1,000$), sentences ($N = 1,000$), and tokens ($N = 10,000$), and varied the window size ($50$ and $100$ tokens), calculating the average MATTR for each combination of factors. 
We found that the ratios were effectively unaffected by the sampling dimension and were largely consistent when varying window sizes, with a bootstrapped MATTR of $.54$ for Photobook, $.65$ for MeetUp! and $.63$ for our data when the window size is $50$, and $.77$, $.68$, and $.75$, respectively, when the window size is $100$ (all $SD$s $\le.02$, rounded to the nearest hundredths). We did observe a difference between the datasets, with MeetUp! averaging a slightly higher MATTR than the data from our task, but both scoring noticeably higher than PhotoBook, suggesting a higher degree of lexical diversity for the former two than the latter one.

As an additional point of comparison, we computed the MATTR for a task that restricts the content of the messages, namely \textit{GuessWhat?!} \cite{DeVries2017GuessWhatDialogue}. The MATTR for GuessWhat?! was $.45$ for a window size of $50$ and $.35$ for a window size of $100$. %

\subsection{Ratio of Contributions} %
To gauge to what extent participants actively contribute to the discourse, we started by comparing the number of messages exchanged between each pair of players over the course of their interaction. We would expect a roughly equal number of turns from each player for interactions in which contributions are balanced and initiative mixed. %
This ratio is calculated as the maximum of the number of messages sent as a proportion of the total number of messages sent.
Expressed as a decimal, a value %
close to $.50$ means participants have sent a (near) equal number of messages over the course of their interaction. The average ratio %
over all interactions was $.52$ $(SD=.01)$, indicating that, overall, participants contributed roughly equally to the discourse in terms of the number of messages exchanged. The ratios for MeetUp! and PhotoBook were $.60$ $(SD=.08)$ and $.53$ $(SD=.03)$, respectively.

However, even with a roughly equal number of messages exchanged it is possible that one player is more proactive while the other is more reactive. In order to measure the extent to which the task results in mixed-initiative dialogue, we calculated a ratio similar to that of the aforementioned contributions, but focused on the proportion of first mentions instead, i.e., the maximum of the number of first mentions as a proportion of the total number of first mentions. We counted for each player, for each round, the number of times they were the first to refer to any of the images. We assume that players that are more actively engaged with the task are more likely to take initiative and proactively make proposals leading to a higher number of first mentions.
For first mentions, the average ratio %
over all interactions was $.60$ $(SD=.07)$, meaning the task tended to skew slightly to one player taking on a more proactive role, but can still be said to have led to mixed-initiative dialogue overall.

\begin{figure}[!t]
\begin{center}
\includegraphics[clip,trim={0.2cm 0.45cm 0.5cm 0.4cm},scale=.47]{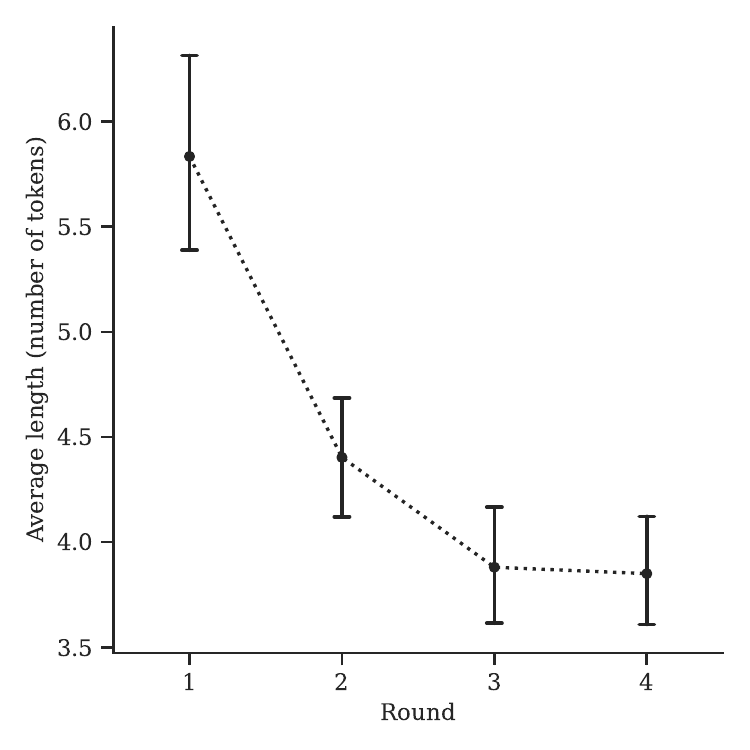} 
\caption{Average length (number of tokens) of referring expressions per round. Graph indicates central tendency trend over the course of the interaction. Error bars show 95\% bootstrapped confidence intervals.}
\label{figure-refexpslength}
\end{center}
\end{figure}

\subsection{Referring Expressions}
In order to come to understand how participants produce and ground referring expressions over the course of an interaction, we resorted to manually annotating all referring expressions in the dataset. In this process the self-annotations, even when noisy, help resolve possible ambiguities. To study how the average length of the referring expressions changes over time, we counted the number of tokens for mentions that refer to one or more images, but marked only those expressions that can be said to refer to the image itself. This means that, aside from generic references, this also excluded modifiers %
that appeared in subsequent utterances. %
As can be seen from Figure \ref{figure-refexpslength}, when plotting the average numbers per round over all interactions we found that referring expressions were noticeably longer in the first round compared to the last (for this calculation we excluded pronouns and noun phrases without content words, e.g., \textit{``the last one''}). The trend that emerged hinted at participants refining their referring expressions, compressing the descriptions over the course of the interaction, and ultimately forming conceptual pacts.

For the calculations that follow, we considered all referential noun phrases, including pronouns and elliptical constructions.
In addition to the possibility of referring expressions referencing multiple images, utterances may contain more than one mention. 
We found that about 17 percent of all utterances contained two or more referring expressions%
, with varying combinations of references to singular images or descriptions of sets of images. %
Almost 60 percent %
of all messages contained referring expressions that can be said to target one or more of the images directly. 
Messages without such expressions may still contain some referring language, such as bridges, %
but we did not consider those to be independent mentions for this calculation. 
We found that just under 30 percent of all utterances %
that did contain referring expressions, contained two or more. %
Furthermore, roughly 10 percent of all referring expressions %
were references that grouped together multiple images under a single description. %

As an indicator of how frequently conversational partners produced referring expressions that were either ambiguous or for which the ambiguity was not resolved prior to the participants proceeding with locking images, we can use the number of times participants locked in different images following an apparent agreement over which image to lock. 
The average frequency with which these confirmed misalignments %
occurred over all interactions was $5.27$ ($SD=3.93$). This means that, on average, both participants locked in a different image more than five times over the course of their interaction, suggesting that referring expressions were not infrequently underspecified %
while participants assumed they were in agreement over which image was being discussed.
An example of this is an interaction in which a speaker simply described a dog as \textit{``the older one''} despite the modifier \textit{``old''} having previously been used only to refer to a different dog than the one intended by the speaker. As a result, the addressee, assuming a mutual understanding, i.e., a pact, had formed around the use of this term for the image that was initially referred to as \textit{``old''} locked in a different image than the speaker. %

We also found various forms of overspecification and negations, both of which are illustrated by the following exchange: 
A: \textit{``then the black one without round cream?''}; 
B: \textit{``do you mean the one with almond topping and chocolate?''}; 
A: \textit{``yes and without a fork''}. 
The first message from participant A is a noteworthy mix of underspecified and overspecified information, as it contains the modifier %
\textit{``without round cream''}, despite no image that is left unranked on the grid containing what is considered by the participants to be \textit{``round cream''}. We found, however, that the image that was ranked and discussed just prior to have been referred to as \textit{``the one with round cream''}. The phrase \textit{``without round cream''} as well as \textit{``without a fork''} in the second message sent by participant A are examples of negations, where the participant draws attention to a dissimilarity between images that focuses explicitly on content that is not present in the referent, but that is visible in the distractors. %
It should be noted that this exchange also exemplifies the collaborative referential process. In addition, the referring expression \textit{``and now the one you mentioned''}, taken from the same interaction, %
demonstrates the need for discourse memory, as without knowledge of the preceding dialogue the phrase is impossible to ground.

\subsection{Ranking Strategies}
\begin{figure}[!t]
\begin{center}
\includegraphics[clip,trim={0.35cm 0.5cm 0.2cm 0.7cm},scale=0.42]{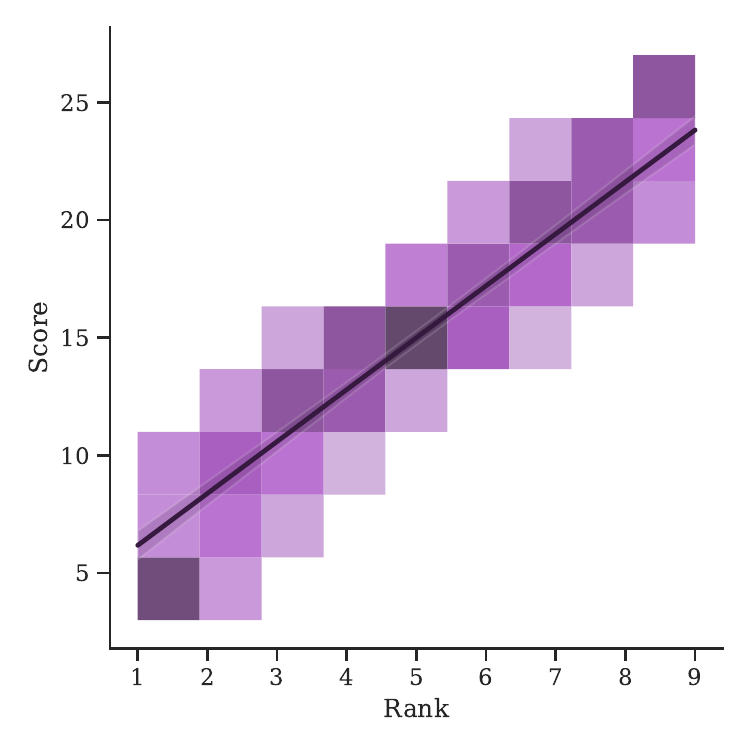} 
\caption{Bivariate histogram showing the distribution of image ranks as sums of scores for each scenario. %
Line indicates linear best fit. Error band shows 95\% bootstrapped confidence interval.}
\label{figure-consensus}
\end{center}
\end{figure}

To assess the extent to which independent pairs of players reached similar agreements in terms of the ranks assigned to images for a given scenario, we converted the ranks to scores. For each image the score is simply the rank assigned to it by the participants; as the gameboard consisted of nine images, the score for the highest ranked image was $1$, the score for the second-highest ranked image was $2$, and so on, with the score for the lowest ranked image being $9$. For each scenario, we then summed the scores for each image and sorted the summed scores in ascending order. In our collected data, we have three independent data points for each scenario, meaning that the lowest attainable score for an image was $3$ and the maximum score was $27$. The results of this analysis are shown in Figure \ref{figure-consensus}.
We would expect a uniform or approximately uniform distribution in the event that, on average, the scenarios lead to diverging strategies or arbitrary rankings. Instead, we saw a clear trend emerge where pairs of players independently converged on similar ranking strategies.

\subsection{Dialogue Acts}
Examination of the conversations showed that our task managed to capture a wide variety of dialogue acts, both with and without referring language. Examples include, but are not limited to, openings (e.g., \textit{``Hi there!''}) and closings (e.g., \textit{``im gonna go now. bye!''}), questions of different types including yes-no (e.g., \textit{``shall we pick the abstract one now?''}) as well as wh (e.g., \textit{``What are your thoughts?''}) which also concern clarification requests (e.g., \textit{``do you mean the one with almond topping and chocolate?''}), (motivated) propositions (e.g., \textit{``I think we should choose the black blackberry. I heard that blackberry is good at business stuff like viewing documents.''}), acceptances (e.g., \textit{``Yea, sounds good to me.''}) and (implicit) rejections (e.g., \textit{``I think round ones are better, that one seems like a rectangle''}) although mostly in the form of (motivated) counter-proposals (e.g., A: \textit{``I would either go for the dotted one or the one with a boat in the middle going out from a port''}; B: \textit{``I'd go for the two boats one first. I think kids all like the paintings to be full''}), and even backchannels (e.g., \textit{``hmmm''}).
It should be noted that we did often find multiple acts within a single message. For example, in the utterance \textit{``Ok, I think french bulldog looks to be the most fierce one. Maybe we pick that one first?''}, the message starts with a discourse marker, \textit{``Ok''}, that is followed by an assertive statement \textit{``I think french bulldog looks to be the most fierce one.''}, which leads into a proposition formulated as a yes-no question, \textit{``Maybe we pick that one first?''}.

\section{Discussion} %
With the introduction of \textbf{A Game Of Sorts}, we aimed to provide a challenging resource to aid visually-grounded dialogue modeling and evaluation efforts. 
In order to benchmark the performance of these models using our task, in particular in terms of their referring expression generation and grounding capabilities, we intended for the collected data to be, to the largest possible degree and in spite of experimental constraints, representative of discussions that involve the collaborative referential process.
Accordingly, we expected to observe a variety of dialogue phenomena that are commonly associated with conversations in which the parties involved collaborate to solve a problem grounded in the visual domain. 

From the results of the small-scale data collection experiment presented in Section \ref{section-results}, we can deduce that the task, as described in Section \ref{sectionmethod}, is capable of such a feat. 
Seeing as the task is intended to enable the study of the collaborative referring expression production and grounding process, the fact that referring language use is frequent should come as no surprise, but the data nevertheless shows that conversations mediated by the game are not simply exchanges of referring expressions.
We see that both parties actively contribute to the discourse, resulting in mixed-initiative dialogues. In comparison with related work, we find that dialogues collected with our task are on average longer. The typical trend for TTR is to decrease with an increase in text length as the author 
exhausts their vocabulary and repetition of previously used words becomes increasingly more likely. %
We nevertheless observe that, in spite of the significantly longer conversations, our data scores relatively high in terms of the overall MATTR. %
One could suggest as a possible explanation for this result that our task fails to capture convergent language use that is common to conversation \cite{Garrod1987SayingCo-ordination}, but a clear indicator for this not being the case comes with the compression of referring expressions over the course of the interaction as shown in Figure \ref{figure-refexpslength}. This leads us to conclude that our task is simply more prone to elicit data with a relatively high degree of lexical diversity despite leading to considerably more repeated references than both MeetUp! \cite{Ilinykh2019MeetUpEnvironment} and PhotoBook \cite{Haber2020TheDialogue}.

We find that the task manages to capture the collaborative nature of referring expression production and grounding in dialogue, as we observe various associated phenomena, including, but not limited to, descriptions of referents negotiated over multiple turns with contributions from each participant, the forming of conceptual pacts, self-expansions, repairs, and negations.
We also find that the data includes a large variety of dialogue acts in which these phenomena are embedded.

Despite their subjective connotation, the proposed sorting criteria do not lead to arbitrary rankings, as indicated by Figure \ref{figure-consensus}. The observed distribution of scores reinforces the idea that participation in a game with the proposed scenarios requires the ability to assess whether propositions are reasonable and to make reasonable propositions, as independent pairs of players seem to have arrived at similar ranking strategies. %
This observation is perhaps best illustrated by an example from the dataset. For the mobile phones image category, participants were presented with a scenario in which they were asked to rank images according to how well each mobile phone could work as a hammer. For each of the three interactions in which this scenario was given, the players ended up ranking the same Nokia mobile phone highest. In one interaction, one of the players commented at the start of the round that \textit{``nokia is famous for working in that way''}, with the other player responding \textit{``I know which one you are talking about''} immediately after. Both players proceeded to lock the same image without specifying any further which one of the three Nokia mobile phones they would lock first. This exchange is a clear indication that participants playing our game rely on their world knowledge to reason about the scenarios. %

Although we conclude, based on analysis of the collected data, that the task as proposed is effective in obtaining the type of lexically-diverse, mixed-initiative dialogues that we sought, we leave verification of whether our observations hold when the game is deployed and data is collected at a larger scale for future work.
Similarly, establishing formal benchmark and evaluation procedures for estimating end-to-end performance on this task merits a dedicated effort.
Aside from additional data collection experiments and formalizing end-to-end evaluation, we see several possible avenues to extend the work presented in this paper. More fine-grained annotations of the referring language, both for the collected data presented here, as well as for future datasets collected using our task, such as part annotations that map the words or phrases of referring expressions to the areas in the images to which they refer, would be a useful addition. 
This could be done post-hoc through manual annotation, but when moving from written to spoken dialogue, fine-grained self-annotation using an approach similar to that of Localized Narratives \cite{Pont-Tuset2020ConnectingNarratives} becomes a possibility. This is also likely to result in more efficient communication, as in its written form and with the current means of self-annotation the interaction can be quite demanding, which likely adversely affects the productiveness of the discussions. %
Finally, although the proposed setup is meant for dyadic communication, the task could be configured to allow for the study of the dynamics of multi-party interactions instead. Other factors that could potentially influence conversational dynamics are not so much in the number of dialogue participants, but more in the nature of their identities; running experiments when controlling for, for example, specific demographics in participant selection could lead to insightful results. 

\section{Acknowledgements}
This work was partially supported by the Wallenberg AI, Autonomous Systems and Software Program (WASP) funded by the Knut and Alice Wallenberg Foundation. The authors would like to thank all colleagues, friends, and family that were involved in testing the various iterations of the game, and Chris Emmery, Travis Wiltshire, Chris van der Lee, Bertrand Higy, Ulme Wennberg, Johan Boye, and the anonymous reviewers for their helpful comments.

\section{Bibliographical References}\label{reference}

\bibliographystyle{lrec2022-bib}
\bibliography{references}

\newpage
\section*{Appendix}
\setcounter{figure}{0}

\begin{figure*}[t!]
\begin{center}
\includegraphics[clip,trim={10cm 3cm 5.5cm 0cm},scale=0.3]{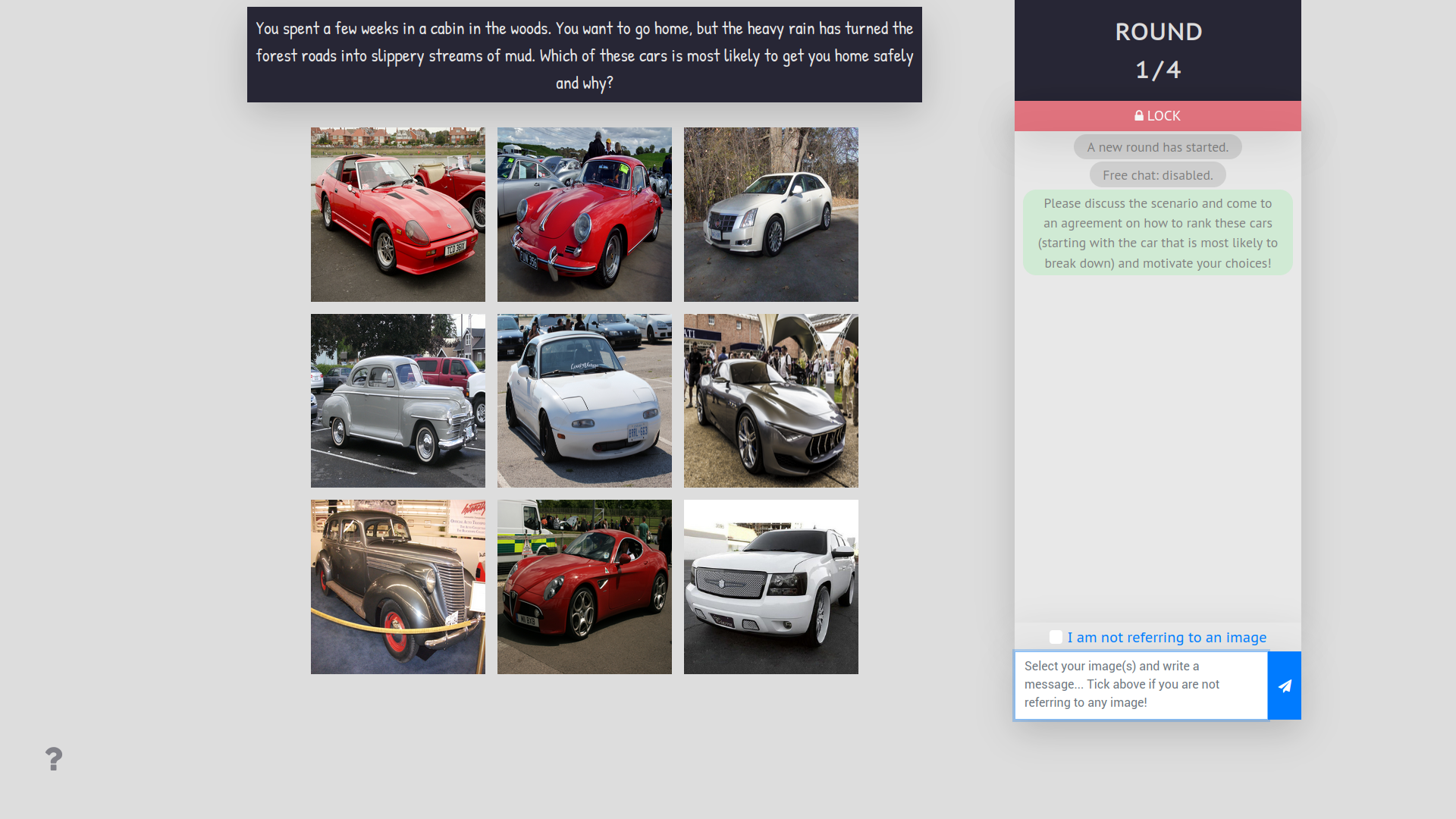} 
\caption{User interface of \textbf{A Game Of Sorts}.}
\label{figure-interface}
\end{center}
\end{figure*}

The following is an excerpt of the first round of game play for the cars image category. Players were presented with the following scenario: \textit{ "You spent a few weeks in a cabin in the woods. You want to go home, but the heavy rain has turned the forest roads into slippery streams of mud. Which of these cars is most likely to get you home safely and why? Please discuss the scenario and come to an agreement on how to rank these cars (starting with the car that is most likely to get you home safely) and motivate your choices!"}.
The highlighted text shows (coarsely) the annotated mentions for this round of dialogue. The images associated with this scenario are shown in Figure \ref{figure-interface}.

\begin{table*}[t!]
\footnotesize
\begin{center}
\begin{tabularx}{\textwidth}{ll}

        Speaker & Message  \\
      \hline
Player A & Hello!	\\
Player B & Hi :)	\\
Player A & Ok, let's get started!	\\
Player B & Okay.	\\
Player A & What do you think?	\\
Player B & I think we should think about the wheels. Do you have any idea which kind of wheels is better to \\&survive from the Mud?	\\
Player A & I think bigger ones are generally better, so that the wheels don't sink in too deep and the car is stuck. \\&By that standard, \hl{the white SUV} might be best option.	\\
Player B & yes I agree that makes sense. But as I am not good at recognising cars and their brands I need more \\&explanations. Do you mean \hl{the one with the black front lights}?	\\
Player A & No problem. Yes, \hl{the one with black front lights and the bottom half of the mirror is also back, and} \\&\hl{the wheels appear to be white}. Do know which one I'm talking about?	\\
Player B & yes, let's lock \hl{it} first!	\\
Player B & $<lock\_event>$	\\
Player A & $<lock\_event>$	\\
Player A & Nice! So I think we should look at how new is the car and the general condition of the car. \\&Newer cars are probably built better. \hl{The white car with silver wheels} seems to be a newer car in \\&good condition.	\\
Player B & New cars should be more capable of getting out of mud than old ones.   let's consider the wheel and \\&\hl{the new modern ones}!	\\
Player B & oh the same! :D	\\
Player A & Agreed. What do you think about \hl{the one I'm talking about, the white one with silver wheels}?	\\
Player B & yes, let's go for \hl{that one}!	\\
Player A & $<lock\_event>$	\\
Player B & $<lock\_event>$	\\
Player A & Ok! I think \hl{the grey car with silver wheels} also looks new. Let's choose \hl{that one} next?	\\
Player B & yes!	\\
Player B & $<lock\_event>$	\\
Player A & $<lock\_event>$	\\
Player B & \hl{the other white one with black wheels and small lights} looks modern and well designed!	\\
Player A & Ok, yea, let's go for \hl{that one}.	\\
Player A & $<lock\_event>$	\\
Player B & $<lock\_event>$	\\
Player A & Great. Then \hl{the red car with exactly 5 circular patterns} could be next? \hl{It} also looks quite new.	\\
Player A & \hl{5 circular patterns in the wheels}.	\\
Player B & yes, I was also thinking about \hl{that one}.	\\
Player B & $<lock\_event>$	\\
Player A & $<lock\_event>$	\\
Player A & Ok! How about \hl{the red car with open top} next?	\\
Player A & \hl{The one that looks less old}.	\\
Player B & yes, \hl{the red one without the roof} looks newer than \hl{the others}!	\\
Player A & $<lock\_event>$	\\
Player B & $<lock\_event>$	\\
Player B & Now it is hard for me to guess which one is older	\\
Player A & \hl{The rest} are all old cars. But it looks like \hl{the silver one} is still being driven while \hl{the grey one} is \\&in the museum. Let's choose \hl{the silver one}, then \hl{the red one}, then \hl{the grey one}?	\\
Player B & yes! great reasoning I didn't pay attention to the surroundings! let's follow your plan!	\\
Player B & $<lock\_event>$	\\
Player A & $<lock\_event>$	\\
Player A & $<lock\_event>$	\\
Player B & $<lock\_event>$	\\
Player A & $<lock\_event>$	\\
Player B & $<lock\_event>$	\\
Player A & Nice!	\\
Player A & Next round?	\\
Player B & :)	\\
Player B & yess	\\

      \hline

\end{tabularx}
\label{table-excerpt}
 \end{center}
\end{table*}

\end{document}